\documentclass{mva_style}
\usepackage{glossaries}
\usepackage[inline]{enumitem}
\usepackage{url}
\makeglossaries
\usepackage{graphicx}

\finalcopy 

\newacronym{fn}{FN}{false negative}
\newacronym{fp}{FP}{false positive}
\newacronym{tn}{TN}{true negative}
\newacronym{tp}{TP}{true positive}

\newacronym{il}{IL}{interest level}
\newacronym{ptz}{PTZ}{pan-tilt-zoom}
\newacronym{hog}{HOG}{Histogram of Gradients}
\newacronym{dpm}{DPM}{Deformable Parts Model}
\newacronym{acf}{ACF}{Aggregate Channel Features}
\newacronym{bgs}{BGS}{background subtraction}

\begin{document}
\title{The autonomous hidden camera crew}

\author{
  Timothy Callemein\thanks{Timothy Callemein and Wiebe Van Ranst contributed equally to this paper.}\\
  {\tt timothy.callemein@kuleuven.be}\\
  \and
  Wiebe Van Ranst\footnotemark[1]\\
  {\tt wiebe.vanranst@kuleuven.be}\\
  \and
  Toon Goedem\'{e}\\
  {\tt toon.goedeme@kuleuven.be}\\  
  EAVISE, KU LEUVEN, BELGIUM\\
}

\maketitle

\section*{\centering Abstract}

\textit{
Reality TV shows that follow people in their day-to-day lives are not a new concept. However, the traditional methods used in the industry require a lot of manual labour and need the presence of at least one physical camera man. Because of this, the subjects tend to behave differently when they are aware of being recorded.
}

\textit{
This paper will present an approach to follow people in their day-to-day lives, for long periods of time (months to years), while being as unobtrusive as possible. To do this, we use unmanned cinematographically-aware cameras hidden in people's houses. 
Our contribution in this paper is twofold: First, we create a system to limit the amount of recorded data by intelligently controlling a video switch matrix, in combination with a multi-channel recorder. Second, we create a virtual camera man by controlling a PTZ camera to automatically make cinematographically pleasing shots.
}

\textit{
Throughout this paper, we worked closely with a real camera crew. This enabled us to compare the results of our system to the work of trained professionals.
}

\section{Introduction}\label{sec:intro}

Disturbing events occur in each household, both negative such as death, divorce, and disease, as well as positive ones like birth, adoption, and marriage.
An independent video production house wants to make a reality TV show that will get families on TV in these situations, hoping to yield emotional and socially inspiring footage. 
Take for example first-time parents questioning how other couples manage to get up at night to care for their newborn child.

Recording a household with cameras is not groundbreaking in itself, various TV programs have 
already explored this concept.
However, the production house desired an innovative system that was able to automatically record using on side hardware without the presence of a human camera crew. In this paper, we describe the solution we developed for this, composed of a system to decide which camera will be recording (section ~\ref{sec:cam_select}), and a controller for the PTZ to take cinematographic medium shots (as described in section~\ref{sec:ptz}). Remark that the video streams produced by our system form the input for a human film editor. In section~\ref{sec:results} we present the results, both quantitatively and qualitatively by validating it using an experienced camera crew.

\section{Related work}

There are different camera network techniques available like~\cite{liao2005blocation, song2010tracking}, that mainly focus on security applications, mostly involving far away cameras capturing an overview of an environment. In this paper however, cameras will follow cinematographic rules that will be positioned at eye height and fairly close to what they are capturing. The presented system will decide on the spot whether or not to record, making it critical to be fast and reliable, as opposed to surveillance video that is permanently recorded and stored for future analysis. Other techniques like~\cite{daniyal2011multi, doubek2004cinematographic} that follow cinematographic rules edit the video streams  to produce one video montage or only focus on one room. Our system will select multiple cameras placed in multiple rooms and record them minimizing data collection providing only useful material that still needs manual editing. This relatively new research topic is being actively researched~\cite{chen2014autonomous}.

Automatically controlling a \gls{ptz} camera to follow people is not a new concept. In~\cite{hulens2014} a system is proposed for lecture recording that also takes into account some cinematographic rules. The lecturer is tracked in the \gls{ptz} video (no overview camera is used) using a combination of a detector and a tracker, they use a PID loop to continuously keep the lecturer on a fixed position in the frame. While such a system works well for lecture recording, recording people in their house is much more challenging. Our system is able to create shots of multiple people, conform to different cinematographic rules and take pleasing static shots.

\section{Hardware setup}\label{sec:hardware_setup}

\begin{figure}
    \centering
    \includegraphics[width=0.35\textwidth]{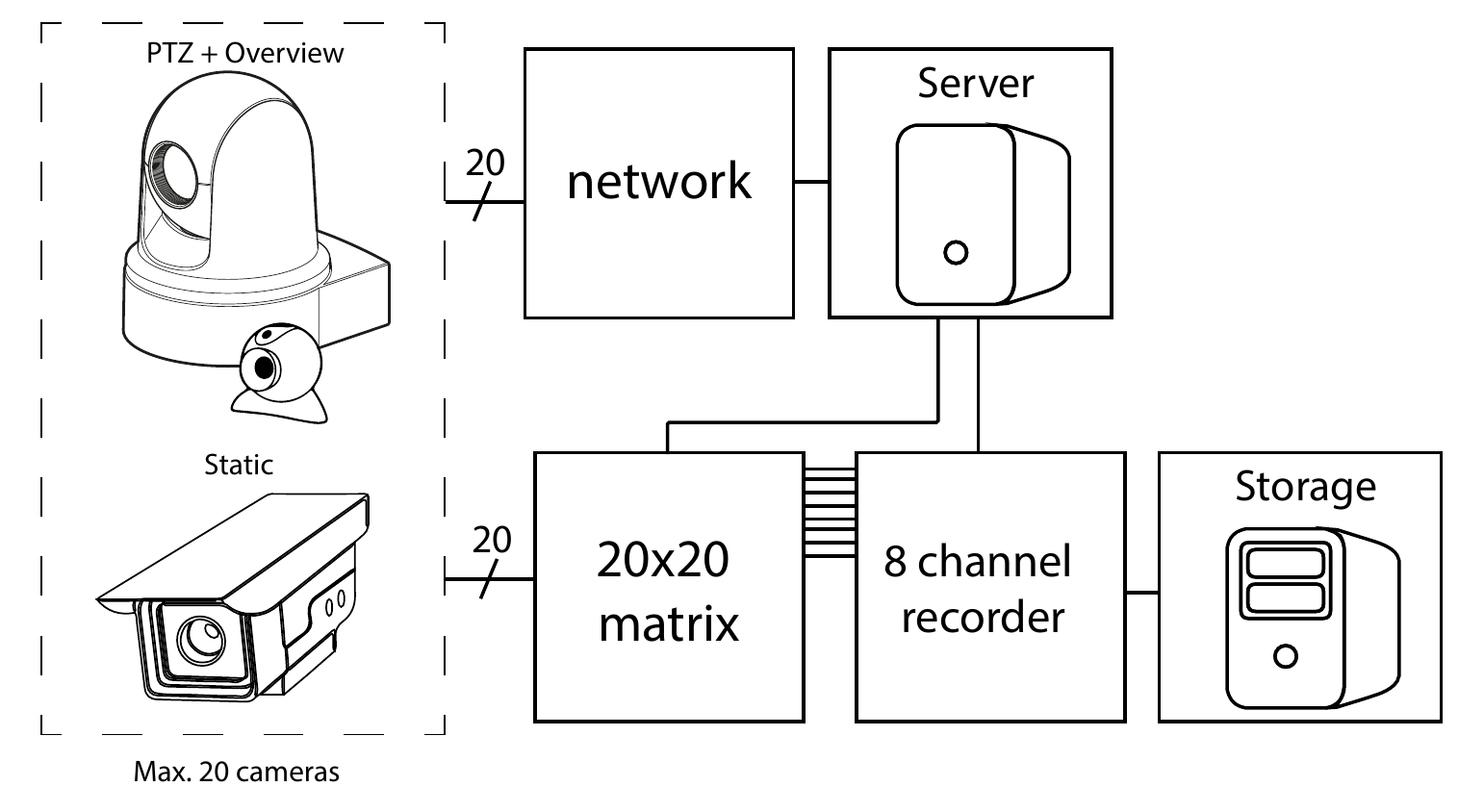}
    \vspace{-2mm}
    \caption{Hardware scheme}
    \label{fig:hardware_schema}
\end{figure}

Figure \ref{fig:hardware_schema} illustrates the hardware setup comprised of multiple Full HD cameras to be recorded. Although the recorder is capable of capturing eight channels simultaneously, we expanded this number by adding a matrix allowing up to 20 cameras in total.

The chosen ProRes LT 422 codec captures at an average of 102 Mbps, amounting up to 1 TB per continuous camera stream per day.  The proposed camera selection method will reduce the needed storage, extending the recording time greatly, and reducing the search space of the human editor.

Two types of cameras are used, static and PTZ cameras, both with a dedicated SDI recording interface and IP interface for image processing and PTZ control. We add an overview camera to each PTZ, to detect actors and steer the PTZ as described in section \ref{sec:ptz}. 

We develop a semi-autonomous calibration procedure to calibrate the PTZ control parameters to the field-of-view of the overview camera. After a rough manual calibration, for each position SURF feature matching is used to correct the local parallax error and to fine align the FoV of the PTZ and the desired canvas cutout from the overview camera.

Our server processes the 20 camera streams in low-resolution through the network, allowing image based decisions that will select the cameras to record (described in section \ref{sec:cam_select}) and control the PTZ (section \ref{sec:ptz}).

\section{Camera selection}\label{sec:cam_select}

One of the camera crew's tasks is deciding when to record, reducing large amounts of useless video material. This requires an algorithm that is able to anticipate the whereabouts of the residents, and their movements between rooms.

To make the record decision automatically, we need to understand what a camera crew uses as criteria for recording cinematographic scenes. 
In a reality TV show that follows the residents, the presence of a person is the primary trigger. In most cases, a script is used to determine when an actor is on set. In reality TV this is not possible, we have to use the entering or exiting of a resident as cue to start recording, which are shots that are also usable in editing. Movie professionals therefore prefer to have a pre-roll: shortly before a subject enters a room, the camera should be switched on.

\begin{figure}
    \centering
    \includegraphics[width=0.3\textwidth]{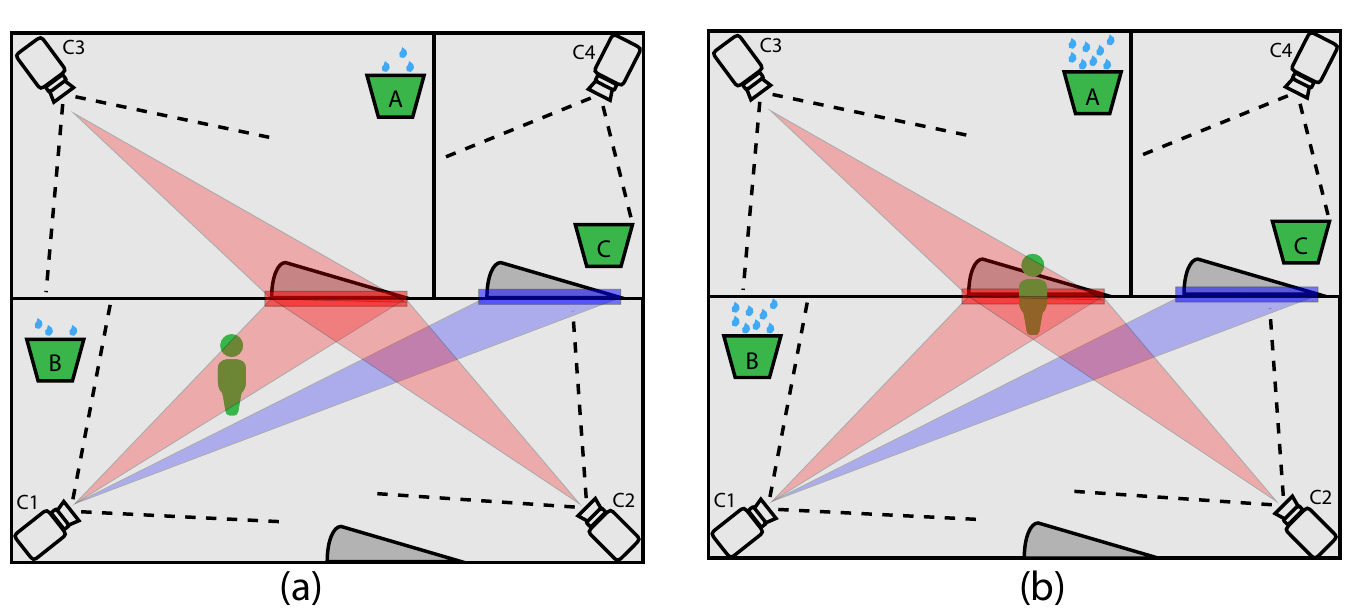}
    \caption{Top view with cameras, buckets and zones adding to the buckets.}
    \vspace{-6mm}
    \label{fig:buckets}
\end{figure}

Figure \ref{fig:buckets} shows a possible setup, three rooms and four partially overlapping cameras. On figure \ref{fig:buckets}a two zones, blue and red, are doors connecting the three rooms. The red zone is visible on  $C_1$, $C_2$, and $C_3$.  Every room also contains a virtual ``bucket'' that holds the room's \gls{il}. The door in red, will raise \gls{il} of the buckets A and B. When a person is present in the red zone of a camera, the connected buckets will fill with water. Then, when a threshold is reached, we start recording the cameras connected to the bucket. This enables cameras that are in a different room, to trigger a room before entering. 

The cameras project the 3D environment on a 2D image, this means a person can be inside a zone, yet far away from a door opening, leading to false positives. Yet, as shown in figure~\ref{fig:buckets}b, multiple cameras will raise the \gls{il} faster, confirming the 3D location from multiple perspectives, providing more speed and reliability when triggering the recording. This combined with different weights per zone, limits false positives and provides fine-tuning capabilities. The buckets also contain a ``hole'', that will lower the \gls{il} over time, greatly reducing the chances of accumulated low activity to trigger a false positive.

We use dedicated static cameras as well as the \gls{ptz} overview cameras to accurately monitor rooms inside a home.  Since both of these cameras are in fixed positions, all movement in the video is caused by changes in the scene, meaning \gls{bgs} techniques work well. 

In all of these cameras' frames we define zones (eg. a door opening) that fill up accompanying buckets. We determine the fill-up rate of a bucket by multiplying the amount of foreground activity in a zone with preset weights that we assigned to each zone. This ensures that eg. a cat walking through a door has less influence than a person.

\section{Autonomous PTZ steering}\label{sec:ptz}
The role of the \gls{ptz} camera in our setup is to replace a traditional camera man without the need for any human intervention.
By adapting the three degrees of freedom of a \gls{ptz} to all kinds of scenarios that take place in front of the camera, we are able make shots that are a lot more interesting than those captured by a static camera. 
To implement such an ``intelligent camera man'' it is essential to have a good understanding of \begin{enumerate*}\item what is currently happening in front of the camera and \item given the knowledge of the current situation how to capture it in a cinematographically pleasing way \end{enumerate*}. 
Note that, contrary to other automated PTZ systems like~\cite{hulens2014} our system's main functionality is to make steady shots. For example, when someone is sitting on a coach watching TV, we do not continuously adjust the camera to keep the, sometimes slightly moving, person at a fixed position in the frame (often determined by the rule of thirds). Cinematographically speaking, such a ``quivering shot'' is not wanted. Instead, we wait for the person to become stationary, calculate what would be a good shot and then when we determine that the time is right, control the \gls{ptz} to make the shot.

\subsection{Understanding the scene}\label{sec:understanding_the_scene}
In our application, the objective of our virtual camera man is to make shots of people. An obvious first step in creating such a system is to find out where people are located in the scene.
To do this we compared different person detectors, like \gls{dpm}~\cite{dubout2012, felzenszwalb2010} or \gls{acf}~\cite{dollar2014} combined with background subtraction and some scene constraints to speed up the algorithm.
For accuracy (especially when dealing with occlusion) and because speed is not our main concern when creating static shots (real-time performance is not necessary) we choose to run a \gls{dpm} upper-body detector. In section~\ref{sec:results} we compare these detectors in more detail on our own dataset.

Apart from a person's location and size, which we get from the person detector, we also need to know the orientation and the position of the head inside the detection. For this, we use the approach proposed by~\cite{hulens2016}. This technique uses the Viola and Jones~\cite{viola2001} face detector with three separate models, one trained on the frontal views, and the other on the left and right profile views of a face dataset. The results are then combined to estimate the gaze of the detected person. Because this method does not rely on small-grained facial features, it is proven to also work on small faces with limited resolution.

Apart from this, we also managed to speed up the detector by combining it with background subtraction. When we detect movement in the foreground image we cut out bounding boxes around active areas. These areas are stored in a pool on which the detector is evaluated. The resulting bounding boxes of the detector (with some added margin) are then added to this pool to be evaluated again (together with the resulting bounding boxes of background subtraction) in the next image. Evaluating places where people were found in the previous image, prevents people who are stationary, ie. invisible to the background subtraction algorithm, from disappearing. 

\subsection{Making cinematographically pleasing shots}\label{sec:making_cinematographically_pleasing_shots}

The detector provides us with the necessary information to produce cinematographic shots: the location, size, position and general gaze direction for each person, detected in the overview camera. We give priority to keeping the maximum amount of people in one shot, over making medium shots with the rule of thirds. This means that the distance between the outer detections will determine the width and the height by the 16:9 ratio of the shots. To provide enough space between the outer detections and the border, we add 15\% to the width. Vertically, we place the estimated eye position of the highest detection to comply with the rule of thirds. 

When there is only one detection found the height and aspect ratio of the detection is used as reference to determine the size of the frame. The position of the eyes is again used to determine the position of the frame on the y-axis. While the general gaze will now determine the position on the x-axis. When a person is looking to the right, we leave space to the right, when he is looking to the left, we provide space to the left. When there is no general gaze available, we choose to place the detection in the center. This is not optimal from a cinematographic perspective, but results in more usable shots than choosing one side as default.
Some examples of the resulting shots are illustrated in figure~\ref{fig:ptz_results}.

\subsection{Switching between different shots}\label{sec:switching_between_different_shots}
As mentioned before, in our application, a continuously moving ``quivering'' camera is not desirable, we want to make steady shots.
To make sure our camera makes steady shots and to prevent it from constantly alternating between different shots, we adjust the camera only if following conditions are met:
\begin{itemize}
	\item The proposed shot is considerably different (determined by position and size) from the current.
	\item The new shot is steady, meaning people in it are only moving slightly or not at all.
    \item The old shot was kept a bit longer ($2\text{s}$ in our case), to allow subjects to leave the scene, and has a duration of at least 6s.
\end{itemize}

If all these conditions are met, we map the proposed canvas from overview camera coordinates to \gls{ptz} control parameters.

\section{Experimental results}\label{sec:results}
In this section, we will report on quantitative results of the three main components of the proposed system (the camera selection, person detector and the cinematographical canvas selection method). However, as the proposed system is currently being used in a grand scale reality TV video production (8 families filmed 24/7 over 12 months), the very evaluation of the system will be in the success of that TV show.

\subsection{Camera Selection}
\begin{figure}
    \centering
    \includegraphics[width=\columnwidth]{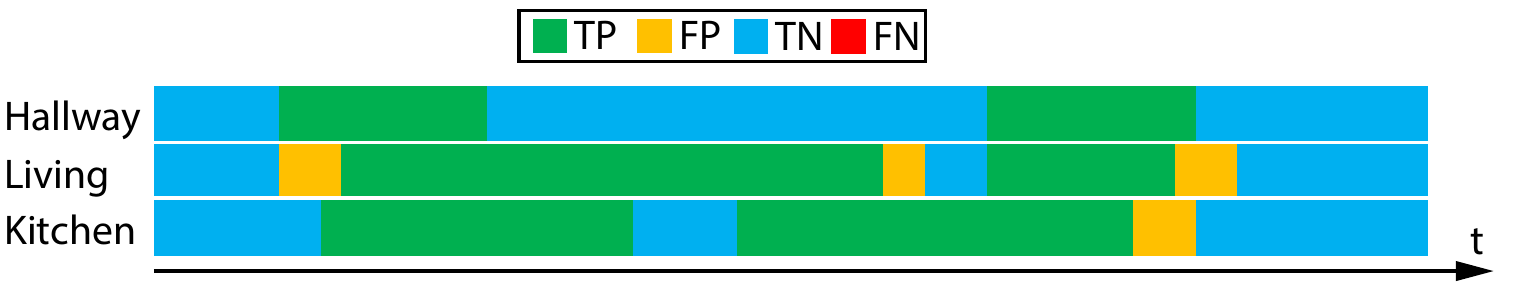}
    \caption{Camera selection results of three buckets during a 10 minute scenario.} 
    \label{fig:cam_sel_results}
\end{figure}

To test the camera selection algorithm we played a scenario inside the test environment, which included movements between different rooms. Figure \ref{fig:cam_sel_results} shows the analysis of three buckets being triggered, triggering recordings, during the 10 minute scenario, which is sampled every 10s and analyzed by comparing the actual output with the expected recording, as given by a real camera man. Since it's important to capture everything (and not lose any action) we set the sensitivity quite high, explaining the presence of multiple \glspl{fp} (triggering while no activity) and also the lack of \glspl{fn} (activity but no or to late triggering of recordings). These results show that the buckets trigger the connected cameras on time, while not missing any vital video information. The accuracy of the tests by room are: hallway 100\%, living room 87\% and kitchen 95\%. In this experiment, we saved 43.33\% of disk space opposed to recording the full 10 minutes and recorded only 8\% data overhead that contains no usable video data.

\subsection{Person detector}
\begin{figure}
    \centering
    \includegraphics[width=0.8\columnwidth]{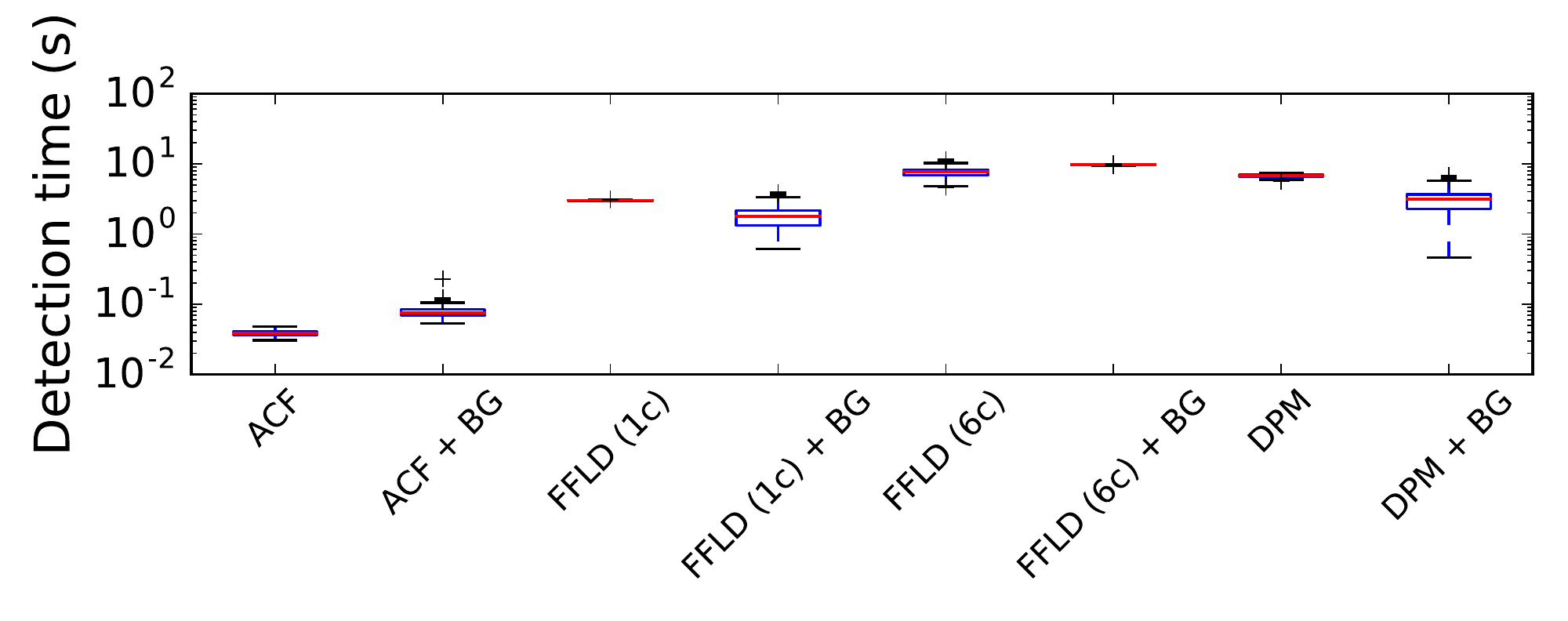}
    \vspace{-2mm}
    \caption{Comparison of the time complexity of different detection algorithms on our test footage.\vspace{-10mm}} 
    \label{fig:det_comp_chart}
\end{figure}
In figure~\ref{fig:det_comp_chart} we compare the time complexity of the different upper body detection algorithms on a test dataset that was recorded in a real setting (inside a typical home). We compare the algorithm with background subtraction enabled (``+ BG'' in figure~\ref{fig:det_comp_chart}). While using background subtraction has a much faster average execution time, there are cases where the entire image needs to be searched causing a big performance hit, for instance when a person walks in front of the camera.

\begin{figure}
    \begin{center}
        \includegraphics[width=0.75\columnwidth]{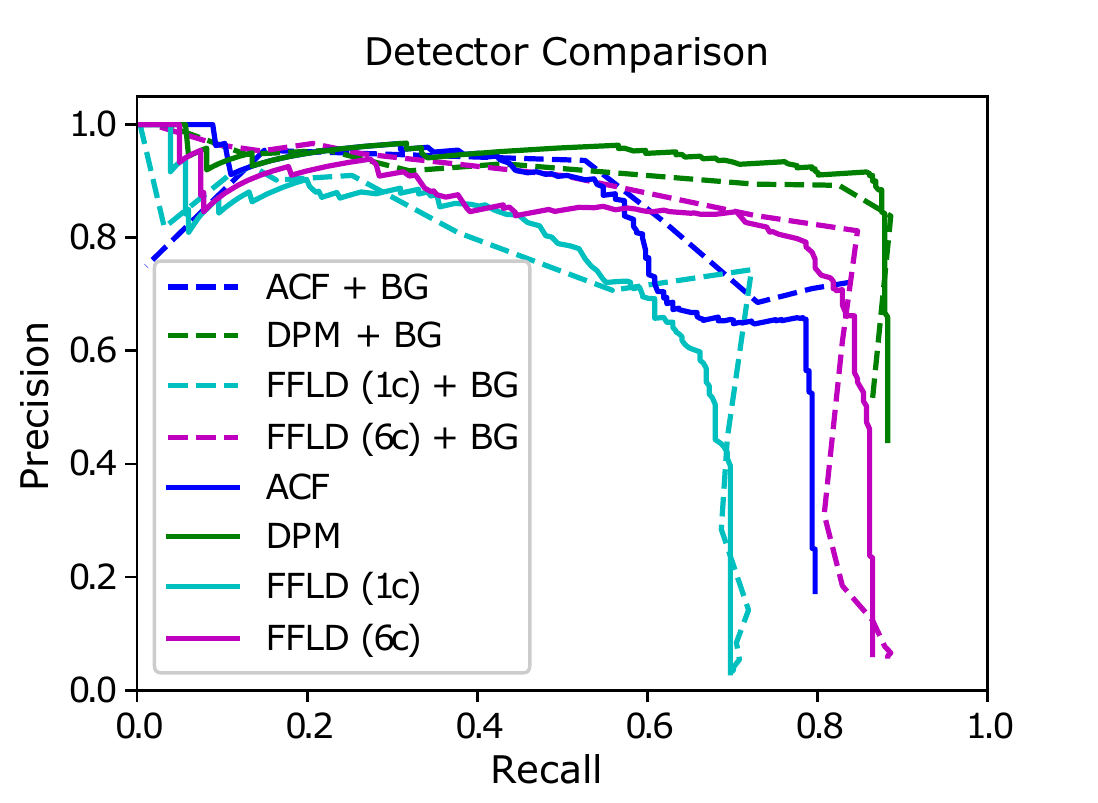}
        \vspace{-2mm}
    \end{center}
    \caption{Detection performance on overview footage from a typical home.}
    \label{fig:det_perf_chart}
\end{figure}

Figure~\ref{fig:det_perf_chart} compares the performance of these detectors on the same dataset. We annotated a total of 105 frames with a total video duration of about two minutes. Some samples of this dataset can also be seen in figure~\ref{fig:ptz_results}. It is clear that the \gls{dpm} upper body detectors (\verb$ffld_6c$ and \verb$dpm$)~\cite{felzenszwalb2010,dubout2012} have the best performance (at the cost of being more computationally intensive), compared to the rigid \gls{acf}~\cite{dollar2014} detector which is not as accurate but much faster.
Limiting the search space by using background subtraction seems to have little impact on detector performance. Adding more people to the scene will only impact detection speed.

\subsection{PTZ canvas} 

\begin{figure}
  \centering
  \includegraphics[width=0.9\columnwidth]{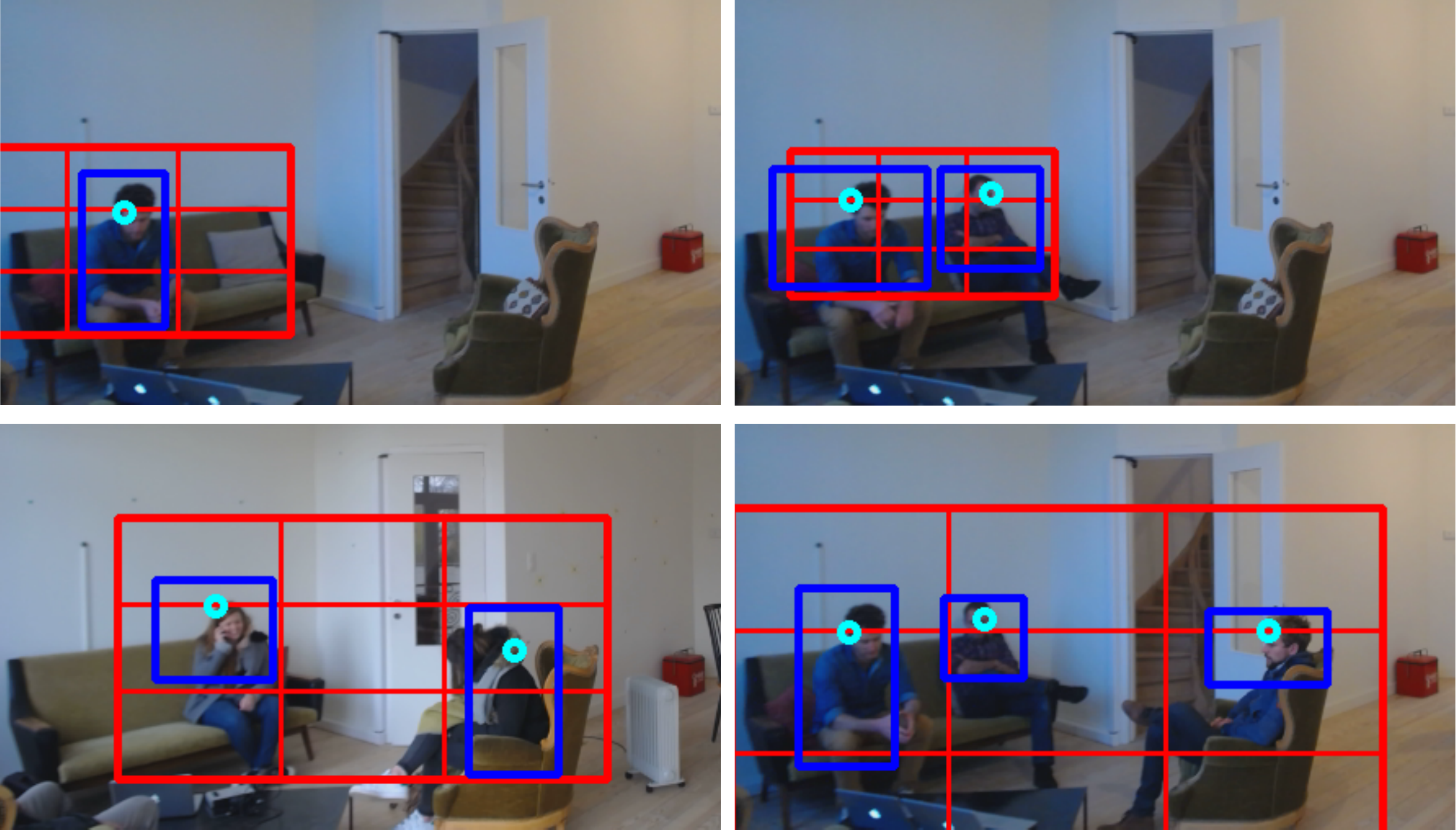}
  \caption{Proposed canvases (red) based on people detections (blue) controlling the \gls{ptz}. Video illustration on http://goo.gl/WIS0P4.}
  \label{fig:ptz_results}
\end{figure}

Figure~\ref{fig:ptz_results} shows the overview image with the detections and the resulting red frame that will be used to control the \gls{ptz}. As can be seen, the resulting canvases are designed to be cinematographically pleasing. To ensure this, we repeated  the same 10 minute scenario 10 times, and continuously let the resulting canvases evaluate by a group of trained camera men. Some video material of these tests can be seen at http://goo.gl/WIS0P4, containing video footage that is selected to be recorded (the multicam video) and multiple videos with detections and proposed canvases. 

\section{Conclusion}\label{sec:conclusion}
In this paper, we proposed a system to automatically film indoor scenarios with multiple static and PTZ cameras, while complying to cinematographic rules. The first task of an autonomous camera crew is choosing whether to record or not. The test results show that our proposed bucket system is able to record all useful video streams with little overhead, decreasing the needed recording storage and search space for the human editor. The next task was using a \gls{ptz} that is able to take medium shots of people detected in an overview camera. Results show that the \gls{ptz} is taking multiple cinematographically pleasing static shots of single and multiple persons based on the overview camera info. Remark that the system we have developed is currently being deployed in a real reality TV video production, providing us with more data to be validated.

\section*{Acknowledgments}
This work is supported by video production house Geronimo and the KU Leuven Cametron project.

\bibliographystyle{siam}
\bibliography{bib} 

\end{document}